\documentclass{article}

 \usepackage[main, final]{neurips_2025}

\usepackage[utf8]{inputenc} 
\usepackage[T1]{fontenc}    
\usepackage{hyperref}       
\usepackage{url}            
\usepackage{booktabs}       
\usepackage{amsfonts}       
\usepackage{nicefrac}       
\usepackage{microtype}      
\usepackage{xcolor} 
\usepackage{graphicx}
\usepackage{comment}
\usepackage{wrapfig}
\usepackage{subcaption}
\usepackage[most]{tcolorbox}

\title{Cross-Disciplinary Knowledge Retrieval and Synthesis: A Compound AI Architecture for Scientific Discovery}

\author{%
 Svitlana Volkova, Peter Bautista, Avinash Hiriyanna, Gabriel Ganberg,\\ Isabel Erickson, Zachary Klinefelter, Nick Abele, Hsien-Te Kao, Grant Engberson \\
  Aptima, Inc.\\
  Woburn, MA 01801 \\
}

\begin{document}

\maketitle

\begin{abstract}
The exponential growth of scientific knowledge has created significant barriers to cross-disciplinary knowledge discovery, synthesis and research collaboration. In response to this challenge, we present BioSage, a novel compound AI architecture that integrates Large Language Models (LLMs) with Retrieval Augmented Generation (RAG), orchestrated specialized agents and tools to enable discoveries across AI, data science, biomedical, and biosecurity domains. Our system features several specialized agents including the retrieval agent with query planning and response synthesis that enable knowledge retrieval across domains with citation-backed responses, cross-disciplinary translation agents that align specialized terminology and methodologies, and reasoning agents that synthesize domain-specific insights with transparency, traceability and usability. We demonstrate the effectiveness of our BioSage system through a rigorous evaluation on scientific benchmarks (LitQA2, GPQA, WMDP, HLE-Bio) and introduce a new cross-modal benchmark for biology and AI, showing that our BioSage agents outperform vanilla and RAG approaches by 13\%-21\% powered by Llama 3.1. 70B and GPT-4o models. We perform causal investigations into compound AI system behavior and report significant performance improvements by adding RAG and agents over the vanilla models. Unlike other systems, our solution is driven by user-centric design principles and orchestrates specialized user-agent interaction workflows supporting scientific activities including but not limited to summarization, research debate and brainstorming. Our ongoing work focuses on multimodal retrieval and reasoning over charts, tables, and structured scientific data, along with developing comprehensive multimodal benchmarks for cross-disciplinary discovery. Our compound AI solution demonstrates significant potential for accelerating scientific advancement by reducing barriers between traditionally siloed domains.
\end{abstract}

\section{Introduction}
The extreme growth of scientific knowledge~\cite{bornmann2021growth} has transformed modern research into an increasingly challenging landscape for individual researchers to navigate. With over one million new papers published annually, alongside vast repositories of domain-specific resources such as chemical structures, medical imaging libraries, and specialized ontologies, researchers face unprecedented difficulty in synthesizing insights across disciplinary boundaries~\cite{wang2023scientific, park2023papers}.This information explosion has created significant barriers to cross-disciplinary collaboration and knowledge discovery, despite evidence that scientific progress increasingly depends on bridging traditionally siloed domains \cite{guimera2020bayesian}. Recent advances in artificial intelligence have demonstrated potential to address these challenges, with domain-specific applications showing promise in areas such as materials science \cite{gomez2016design} and drug discovery \cite{sadybekov2022synthon}. However, existing approaches primarily focus on information retrieval within siloed domains rather than facilitating meaningful knowledge synthesis across disciplinary boundaries. To address these challenges, we present a novel compound AI system that integrates LLMs, RAG, knowledge graphs (KGs),  specialized agents and tools to enable breakthrough discoveries across AI, computer science and engineering, biomedical, and biosecurity domains --ultimately helping to bridge the communication gap and manage the overwhelming volume of cross-disciplinary declarative (“knowing what”), procedural (“knowing how”) and conditional (“knowing when and why”) scientific  knowledge~\cite{anderson2013architecture,ryle2009concept,squire2004memory,star2013procedural}.

\begin{figure}[htb!]
\begin{center}
\includegraphics[width=10cm]{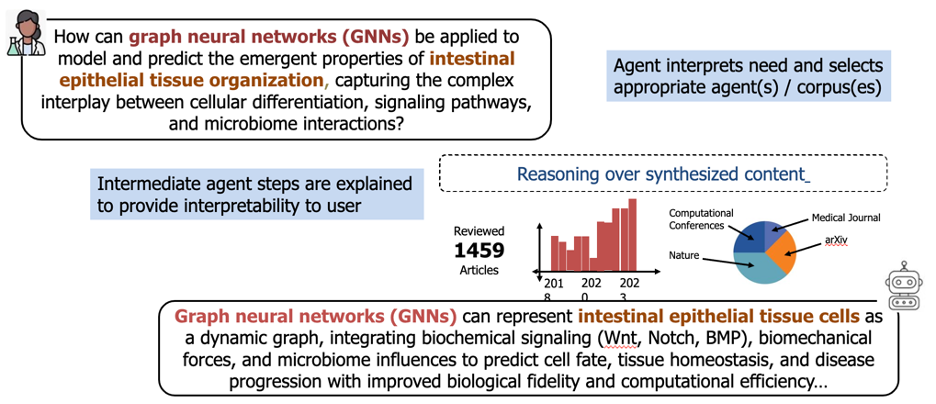}\\
\includegraphics[width=10cm]{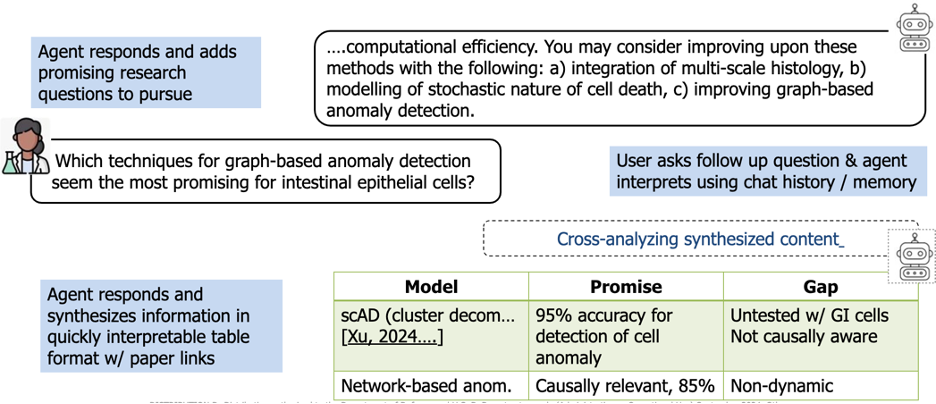}
\end{center}
\caption{User-centric BioSage design demonstrating cross-disciplinary knowledge retrieval and synthesis and contextual conversation flow. The system processes a query, provides transparent explanations of agent reasoning processes while synthesizing knowledge across domains, and maintains conversational context to handle follow-up questions with structured insight presentation to the user.}
\label{fig:frontend}
\end{figure}

User-centric design of BioSage compound AI system enables intuitive and transparent user-agent interactions through an interpretable conversational interface. As shown in Figure~\ref{fig:frontend}, the system incorporates several key features that enhance scientific workflows. When a user poses a cross-disciplinary question, the system selects specialized agents and knowledge corpora to address the query. Throughout the interaction, intermediate agent steps are explicitly explained to users, providing full transparency into the system's reasoning process~\cite{dedhe2023origins,fletcher2012metacognition}. Beyond simply answering queries, the system can represent synthesized information in structured formats that highlight  capabilities, promises, and gaps. The system maintains conversation context through agent memory, allowing it to interpret follow-up questions within the established research thread, thus creating a coherent and productive dialogue.

In this paper, we focus on the implementation and rigorous evaluation of our retrieval agents, while integrating prototypes of other types of agents. BioSage includes three types of agents: (1) retrieval agents that discover and synthesize relevant information across disciplinary boundaries, (2)  translation agents that align specialized terminology and methodologies between fields, and (3) reasoning agents that synthesize domain-specific knowledge into coherent cross-disciplinary insights. 
The key contributions of our work outlined below. 
\begin{itemize}
\item A novel compound AI system~\cite{compound-ai-blog} that orchestrates agents with frontier models to enable scientific knowledge discovery and synthesis across siloed domains.
\item Agentic workflows with retrieval agents with query planning and cross- domain synthesis, translation and reasoning agents.
\item Comprehensive evaluation across SOTA science benchmarks (LitQA2~\cite{laurent2024labbenchmeasuringcapabilitieslanguage}, GPQA~\cite{rein2024gpqa}, WMDP~\cite{li2024wmdp}, HLE~\cite{phan2025humanity}) and a novel cross-disciplinary benchmark developed by our team.
\item Causal investigations into what effects compound AI system performance using models, benchmarks and component ablation e.g., Vanilla LLM, RAG, or agents as interventions. 
\item Insights and details for constructing a new benchmark specifically designed to evaluate cross-disciplinary knowledge discovery and synthesis of compound AI systems.
\item Designs for specialized Human-Agent Interaction (HAI) workflows supporting scientific activities including summarization, research debate, and cross-disciplinary brainstorming.
\end{itemize}

\section{Related Work}
\paragraph{LLMs for Scientific Knowledge Retrieval}
Several AI-powered tools have emerged to help researchers navigate the growing volume of scientific literature. Elicit.ai\footnote{https://elicit.com/} offers automated literature review capabilities by retrieving relevant papers and summarizing key findings, enabling researchers to quickly survey a field without manually sifting through hundreds of papers. SciSpace\footnote{ https://scispace.com/} focuses on extracting insights from individual papers through advanced summarization and knowledge mapping techniques, helping researchers understand complex scientific content more efficiently. HasAnyone.com\footnote{https://hasanyone.com/} addresses the verification challenge by enabling binary validation of scientific claims, searching literature for evidence that either supports or contradicts specific assertions with referenced answers. Consensus\footnote{https://consensus.app/} takes a broader approach by aggregating scientific consensus across multiple studies on specific research questions, helping researchers understand the weight of evidence on contested topics. While these tools provide valuable capabilities for literature retrieval, summarization, and verification, they primarily operate within single domains rather than facilitating true cross-disciplinary knowledge synthesis and discovery.

\paragraph{Foundation Models for Science}
Recent surveys have highlighted the emergence of scientific LLMs~\cite{bommasani2021opportunities} as a specialized subclass of language models specifically engineered to facilitate scientific discovery~\cite{zhang2025scientific,ai4science2023impact}.These models are designed to comprehend and generate domain-specific "scientific language" that extends beyond conventional linguistic boundaries. Scientific LLMs have shown particular promise in biological and chemical domains~\cite{horawalavithana2023scitune,horawalavithana2022foundation,munikoti2023atlantic,munikoti2023evaluating,wagle2023empirical}, with applications ranging from processing textual knowledge to modeling small molecules, macromolecular proteins, and genomic sequences~\cite{dollar2022moljet}. Despite their potential, existing research identifies significant challenges in developing truly effective scientific LLMs, including the need for domain-specific architectures, specialized training datasets, and rigorous evaluation methodologies tailored to scientific tasks~\cite{bommasani2023holistic}. 
Recent work has introduced specialized models for scientific literature exploration like OpenScholar, which combines a retrieval system indexing 45 million papers with language model synthesis capabilities~\cite{asai2024openscholar}. Their research demonstrates that even smaller models (OpenScholar-8B) can outperform larger general-purpose models like GPT-4o when equipped with effective retrieval mechanisms and citation verification. The accompanying ScholarsQABench provides the first large-scale multi-domain benchmark for literature search, establishing a valuable evaluation standard for computer science, physics, neuroscience, and biomedicine.

\paragraph{Agentic AI  for Scientific Discovery}
Several domain-specific agentic AI systems have been developed to support scientific discovery. ChemCrow~\cite{m2024augmenting} integrates expert-designed tools with GPT-4 to enhance performance in chemistry tasks across organic synthesis, drug discovery, and materials design. Similarly,~\cite{boiko2023autonomous} presented an autonomous system capable of designing, planning, and performing complex chemistry experiments. These systems demonstrate the potential of integrating LLMs with domain-specific tools and knowledge sources to enable new scientific capabilities. In the biomedical domain,~\cite{gao2024empowering} proposed a framework for AI agents in biomedical research, envisioning collaborative systems that combine human expertise with AI to analyze large datasets and navigate hypothesis spaces.~\cite{lu2024ai} introduced the AI Scientist, a comprehensive framework for fully automated scientific discovery where LLMs independently generate research ideas, write code, run experiments, and produce complete scientific papers. Similarly, Agent Laboratory (Zhang et al., 2024) provides an autonomous framework that handles the entire research pipeline from literature review to experimentation and report writing, reducing research costs by 84\% compared to previous methods while achieving SOTA performance.

Google's AI co-scientist, built on Gemini 2.0, represents another significant advancement with its multi-agent architecture and tournament evolution process for hypothesis generation and refinement~\cite{gottweis2025towards}. This system has shown promising results in biomedical research areas including drug repurposing where it proposed candidates showing tumor inhibition in vitro and novel target discovery for liver fibrosis with validated anti-fibrotic activity. These agentic AI systems represent a significant step toward collaborative human-AI scientific workflows, where AI agents can handle routine tasks while enabling researchers to focus on creative aspects of scientific inquiry. While these systems demonstrate significant advances in domain-specific applications, they primarily operate within single scientific domains rather than bridging multiple disciplines. The most relevant cross-disciplinary work by~\cite{swanson2024virtual} introduces The Virtual Lab, an AI-human collaborative framework consisting of multiple specialized agents working alongside human researchers, but this approach does not specifically address the challenges of cross-disciplinary knowledge synthesis.

\section{Methodology}
\subsection{User-Centric Design and Human-Agent Interaction Workflows}
Our approach to developing BioSage centers on user-centric design principles that prioritize scientific workflow integration~\cite{shneiderman2022human,endsley2017here}. Unlike many AI systems that focus primarily on model performance metrics, we began by examining how scientists actually work and the specific challenges they face in cross-disciplinary research. Through collaborative sessions with I/O psychologists, human factors experts, and AI developers, we identified three critical scientific workflows that our system needed to support. 

\begin{itemize}
\item{\bf Summarization} This workflow addresses the information overload problem by intelligently filtering and consolidating relevant scientific artifacts  e.g., research papers, reports etc. Scientists can rapidly process new information without manually sifting through hundreds of papers, reducing cognitive load while maintaining awareness of emerging research. The system provides concise syntheses with source traceability, enabling users to quickly identify key findings while maintaining scientific rigor~\cite{lala2023paperqa,laurent2024labbenchmeasuringcapabilitieslanguage}.

\item{\bf Research Debate} This workflow  surfaces contradictory viewpoints and critical perspectives, prompting researchers to reflect on potential flaws or alternative interpretations of scientific claims. By structuring conflicting evidence, the system helps prevent confirmation bias and encourages thorough evaluation of competing hypotheses--a process essential for scientific advancement but often difficult to maintain in practice~\cite{gao2024empowering}.

\item{\bf Brainstorming} This workflow generates novel research ideas and hypotheses, sparking creative discussions within interdisciplinary teams. By drawing connections across traditionally siloed domains, the system can suggest unexpected research directions that might otherwise remain unexplored, while providing structured frameworks for evaluating the feasibility and potential impact of these ideas~\cite{gao2024empowering,schmidgall2025agent}.
\end{itemize}

These workflows are implemented through an intuitive conversational interface that maintains transparency throughout the interaction. The system explicitly communicates its reasoning process, enabling users to understand how conclusions were reached and maintain appropriate trust calibration. Our iterative development process revealed that even relatively straightforward AI components can provide meaningful assistance when strategically positioned within existing scientific workflows.

\subsection{Compound AI Architecture: Frontier Models, RAG, Agents and Tools}
\begin{wrapfigure}{r}{0.6\textwidth}
\begin{center}
\includegraphics[width=8cm]{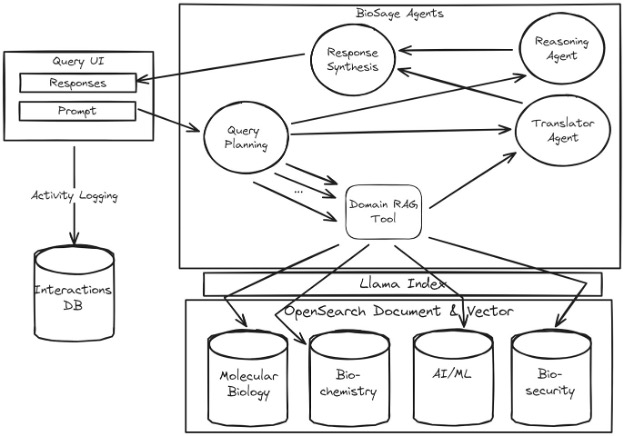}
\end{center}
\caption{BioSage compound AI architecture that integrates specialized agents (Query Planning, Response Synthesis, Reasoning, and Translation) with vectorized domain-specific knowledge bases through LlamaIndex. User queries flow through the UI to the Query Planning agent, which orchestrates workflows across other agents and RAG to access specialized knowledge repositories across domains.}
\label{fig:backend}
\vspace{-0.3cm}
\end{wrapfigure}

BioSage implements a novel compound AI architecture that orchestrates multiple specialized agents and tools within a cohesive system designed for cross-disciplinary scientific discovery. As illustrated in Figure~\ref{fig:backend}, the architecture features a layered design centered around specialized agents that work in concert to process user queries. At the interface level, the Query UI captures user inputs and displays responses while maintaining activity logging for system improvement and user study analysis. The core of the system consists of specialized agents including Retrieval (Query Planning and Response Synthesis), Reasoning, and Translation agents that communicate bidirectionally to collaboratively process complex scientific inquiries described in Appendix.

BioSage's RAG is implemented using LlamaIndex, employing a hybrid semantic search strategy. This method combines traditional term-based filtering with semantic similarity matching, allowing agents to efficiently query individual disciplines and subsets of multiple scientific domains including biology, biochemistry, AI/ML, biosecurity, etc. Research papers and domain-specific information are embedded using the \verb|sentence-transformers/all-MiniLM-L6-v2| model, a lightweight yet effective embedding model that captures semantic meaning across diverse domains. These embeddings are then stored in OpenSearch, providing a scalable and high-performance vector database solution. This approach enables rapid and accurate retrieval of relevant information during query processing.

The BioSage agent system is built on the PydanticAI framework and deployed on AWS r5.2xlarge instance, which provides a structured and type-safe approach to agent development. The Query Planning agent, central to the system's orchestration, utilizes a tool-calling mechanism as a semantic router. This allows it to dynamically select the most appropriate specialized agent for processing each query, ensuring efficient and targeted handling of diverse scientific inquiries.

All user interactions are stored in an Interactions DB to support continuous system improvement through user feedback analysis. This architecture enables BioSage to leverage the strengths of each component—from specialized domain knowledge in the vector stores to the retrieval, translation and reasoning capabilities of the agents—creating a system that transcends the capabilities of any single model while maintaining transparency and trust through explicit communication pathways between components. To ensure responsible deployment and mitigate potential misuse, BioSage implements a safety framework that combines  dual-use risk assessment for scientific queries and automated ethical guardrails that prevent the generation of potentially harmful research directions while preserving scientific utility and discovery capabilities.

\section{Experimental Setup}
We conduct a comprehensive evaluation of BioSage using multiple configurations and benchmarks. Our experiments tested three base models (Llama-3 70B, GPT-4o, and Phi4 14B) across four distinct ablation conditions: LLM without augmentation (Vanilla LLM), RAG with basic retrieval (Vanilla RAG), agents with basic query planning (Agent v1), and agents advanced query planning (Agent v2). Benchmarks include LitQA2, GPQA, WMDP, and HLE-Bio. Each configuration was evaluated on identical test sets to enable direct performance comparisons, with metrics including accuracy and precision. This experimental design allowed us to isolate the impact of each architectural component while validating the overall system effectiveness across knowledge domains.

\paragraph{Cross-Disciplinary RAG Dataset}
In our study, we created a diverse cross-disciplinary dataset by integrating content from S2ORC~\cite{lo2019s2orc} and LitQA2 ~\cite{laurent2024labbenchmeasuringcapabilitieslanguage} collections, comprising 585,902 scientific papers spanning multiple domains. Medicine (187K papers) and Computer Science and Engineering (165K papers) represent the two largest categories, followed by Environmental Science (55K), Biology (42K), and Physics (25K). The "Other" category includes significant contributions from Psychology (15.5K), Chemistry (9.8K), Mathematics (8.2K), and several smaller disciplines. This comprehensive corpus enables the discovery and synthesis of cross-disciplinary knowledge, with articles frequently categorized across multiple domains reflecting the inherently interdisciplinary nature of scientific research. 

\paragraph{SOTA Science Benchmarks}
We evaluate performance of BioSage using several established scientific benchmarks designed for a single domain including LitQA2, GPQA, WMDP, and HLE-Bio. LitQA ~\cite{laurent2024labbenchmeasuringcapabilitieslanguage} and LitQA2 ~\cite{laurent2024labbenchmeasuringcapabilitieslanguage} are designed to assess retrieval-augmented generation capabilities across scientific literature, while GPQA~\cite{rein2024gpqa} offers graduate-level "Google-proof" questions that test deep domain understanding. The WMDP benchmark~\cite{li2024wmdp} measures capabilities while reducing potential for misuse through unlearning techniques. Humanity's Last Exam (HLE)~\cite{phan2025humanity} provides a rigorous assessment of model performance across knowledge domains. 

\paragraph{BioSage Cross-Disciplinary Benchmark}
To evaluate BioSage's cross-disciplinary capabilities effectively, we developed a novel benchmark specifically designed to address the intersection of AI and biomedical disciplines. Our analysis of existing scientific benchmarks revealed a significant gap—while numerous benchmarks evaluate domain-specific capabilities (LitQA2, GPQA, HLE etc.), none adequately measured cross-disciplinary knowledge synthesis and reasoning abilities critical for cross-disciplinary  scientific discovery.

Real-world scientific research increasingly spans traditional domain boundaries, with biomedical advances heavily dependent on AI methodologies. Our compound AI system required evaluation on its core function: retrieving, analyzing, and synthesizing knowledge across both biological content and computational techniques simultaneously. To address this need, we created a specialized benchmark featuring 116 cross-disciplinary question-answer pairs focusing on the intersection of AI with biomedical disciplines. For that, we utilized a zero-shot prompting approach with GPT-o1, generating questions from research papers in biology and AI. After experimenting with multiple methodologies—including full-document prompting, sentence-based chunking, and token-based chunking—we implemented a hybrid chunking with reasoning approach. This method balances semantic integrity with token efficiency while prompting step-by-step reasoning during question generation, producing higher quality questions particularly suitable for evaluating  bio-AI queries. 

The resulting benchmark features multiple-choice questions with detailed reasoning components, such as "Which deep learning models specifically address densely packed nuclei in cellular image segmentation?" and "Which techniques are commonly used in multi-omics data integration for predicting synergistic drug combinations?" These questions require systems to demonstrate understanding across both biological concepts and computational methodologies—precisely the capability BioSage was designed to deliver. Our cross-disciplinary benchmark represents a significant contribution to the field, providing a standardized evaluation framework for cross-disciplinary AI systems and enabling more rigorous assessment of scientific knowledge synthesis capabilities than previously possible with domain-specific benchmarks.

\paragraph{Causal Investigations into Compound AI System Behavior}
Causal analysis allows us to investigate the effects of compound AI system performance by exploring causal linkages and structure (e.g., “when
X is seen to increase, we see a decrease in Y ”). The goal is to move beyond correlation-based analysis by relying on those built around Pearl’s~\cite{pearl2009causality, pearl2018bookofwhy} causal framework. For that, we leverage Structural Equation Modeling (SEM) tools and rely on the NOTEARS algorithm and the package Causalnex \citep{quantumblack2020causalnex, zheng2018notears} to perform causal discovery in the form of directed-acyclic graph (DAG) weights and edges. We focus on treatment/intervention (model type, benchmark etc.) and outcome's (e.g., readability, performance, generation quality etc.) currently, blocking incoming edges from our treatments for logical reasons. Thus, the current experiments  avoid confounding effects where variables that effect outcome also affect treatments at the same time.

\section{Evaluation Results and Analysis}
\begin{figure}[b!]
\centering
\begin{subfigure}{.5\textwidth}
  \centering
  \includegraphics[width=.7\linewidth]{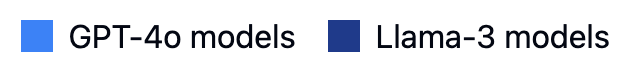}
  \label{fig:legend}
\end{subfigure}\\
\begin{subfigure}{.5\textwidth}
  \centering
  \includegraphics[width=.95\linewidth]{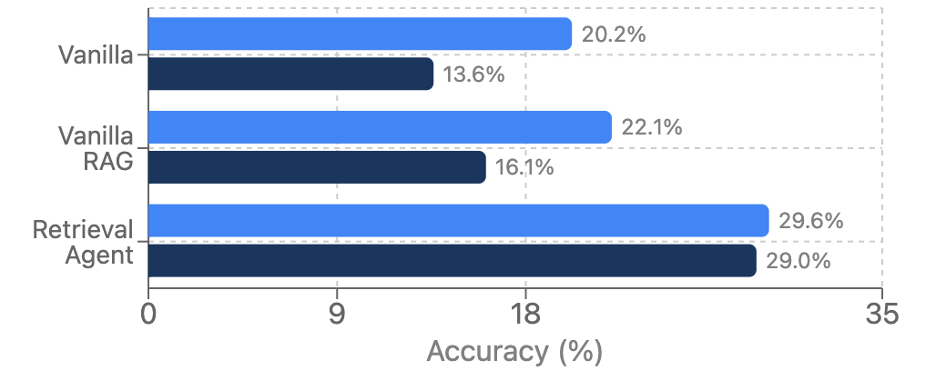}
  \caption{LitQA2~\cite{lala2023paperqa}}
  \label{fig:litqa}
\end{subfigure}%
\begin{subfigure}{.5\textwidth}
  \centering
  \includegraphics[width=.95\linewidth]{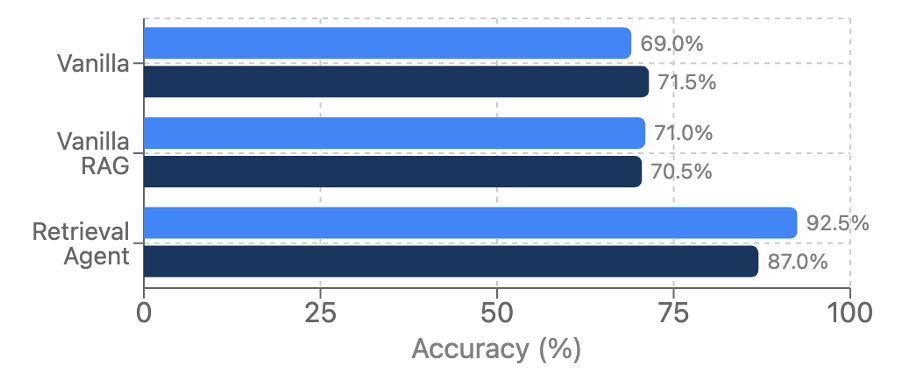}
  \caption{WMDP-200~\cite{li2024wmdp}}
  \label{fig:wmdp}
\end{subfigure}
\begin{subfigure}{.49\textwidth}
  \centering
  \includegraphics[width=1.05\linewidth]{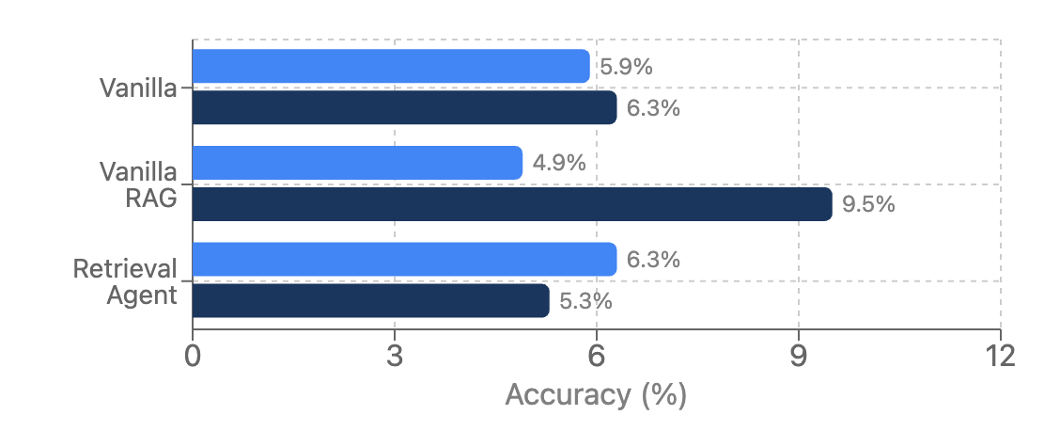}
  \caption{HLE-Bio-303~\cite{phan2025humanity}}
  \label{fig:hle}
\end{subfigure}
\begin{subfigure}{.49\textwidth}
  \centering
  \includegraphics[width=.95\linewidth]{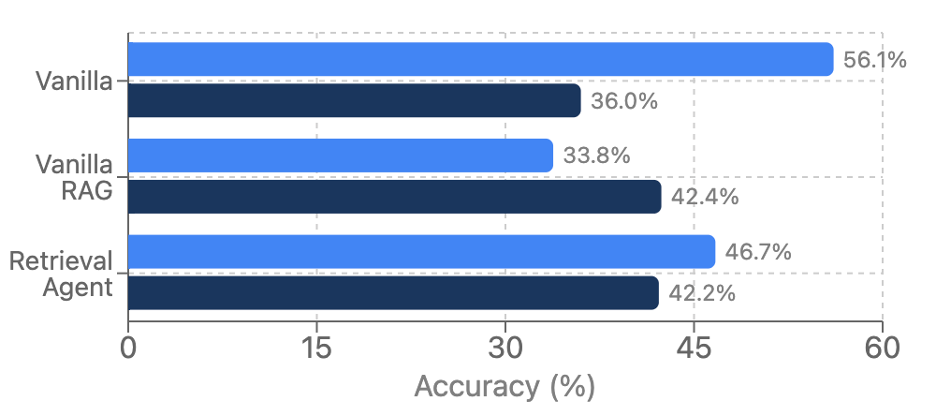}
  \caption{GPQA Diamond~\cite{rein2024gpqa}}
  \label{fig:gpqa}
\end{subfigure}
\caption{Accuracy (\%) results for different BioSage compound AI  configurations on four scientific benchmarks: (a) LitQA2, (b) WMDP with 200 question subset, (c) HLE-Bio with 303 questions, and (d) GPQA Diamond; we compare three configurations: Vanilla LLM (base model), Vanilla RAG, and Retrieval Agent. Blue bars represent GPT-4o model performance, while dark blue bars show Llama-3 model results. The data demonstrates significant performance improvements with BioSage's retrieval agents compared to vanilla configurations, with particularly notable gains on the LitQA2 (29.6\% vs. 20.2\% for GPT-4o) and WMDP (92.5\% vs. 69.0\% for GPT-4o) benchmarks.}
\label{fig:results}
\end{figure}

Our comprehensive evaluation demonstrates that BioSage's compound AI architecture delivers significant performance improvements across multiple scientific benchmarks. We evaluated three foundation models (GPT-4o, Llama-3 70B, and Phi4 14B) in four configurations: vanilla LLM, vanilla RAG, retrieval agent v1, and retrieval agent v2 with cross-domain synthesis capabilities. Results consistently show that our specialized agent architecture enhances model performance on retrieval and reasoning tasks. Figure~\ref{fig:results} demonstrates improvements on the LitQA2 benchmark, where our GPT-4o-powered retrieval agent achieved 29.6\% accuracy—a 46.5\% improvement over the vanilla model baseline (20.2\%)—and on the WMDP benchmark, where the same agent configuration achieved 92.5\% accuracy, representing a dramatic 34.1\% improvement over vanilla GPT-4o (69.0\%). These results validate our approach to cross-disciplinary knowledge synthesis and demonstrate BioSage's ability to effectively bridge traditionally siloed domains.

\paragraph{LitQA2 Evaluation} On the LitQA2 benchmark, which evaluates question-answering abilities over full-text scientific papers, our retrieval agent with query planning and synthesis demonstrated significant performance improvements over baseline configurations. The GPT-4o-powered agent achieved 29.6\% accuracy, representing a 9.4 percentage point (46.5\%) improvement over the vanilla GPT-4o baseline (20.2\%). Similarly, the Llama-3-70B retrieval agent achieved 29.0\% accuracy, a 15.4 percentage point improvement over its vanilla counterpart (13.6\%). These results indicate that our specialized agent architecture effectively enhances model performance on retrieval tasks.

\paragraph{WMDP Evaluation} Results on the WMDP benchmark (200 question subset) revealed even more dramatic improvements. The GPT-4o retrieval agent achieved 92.5\% accuracy, representing a 23.5 percentage point (34.1\%) improvement over vanilla GPT-4o (69.0\%). Similarly, the Llama-3-70B agent achieved 87.0\% accuracy, a 15.5 percentage point improvement over its vanilla counterpart (71.5\%). These substantial gains demonstrate the effectiveness of our agent architecture in enhancing biomedical reasoning capabilities.

\paragraph{HLE-Bio and GPQA Evaluation} On the HLE-Bio benchmark (303 question subset), performance improvements were more modest. The GPT-4o retrieval agent showed a 0.4 percentage point (6.8\%) improvement over vanilla GPT-4o (5.9\% vs 6.3\%), while RAG implementation showed a 1.4 percentage point (28.6\%) improvement (4.9\% vs 6.3\%). For Llama-3-70B, the pattern was different: the retrieval agent performance (5.3\%) was lower than vanilla RAG (9.5\%) but still showed improvement over the vanilla model (6.3\%).

On the GPQA benchmark, we observed that model selection significantly influenced performance. The Llama-3-70B retrieval agent outperformed the vanilla model by 6.2 percentage points (17.2\%), achieving 42.2\% accuracy compared to 36.0\% for the vanilla LLM. However, for GPT-4o, the retrieval agent (46.7\%) underperformed compared to the vanilla model (56.1\%), suggesting that the base model's inherent knowledge was more valuable than our retrieval mechanisms for this benchmark.

\paragraph{Cross-Disciplinary Benchmark Evaluation}
\begin{wrapfigure}{r}{0.6\textwidth}
\vspace{-0.5cm}
\begin{center}
\includegraphics[width=8.5cm]{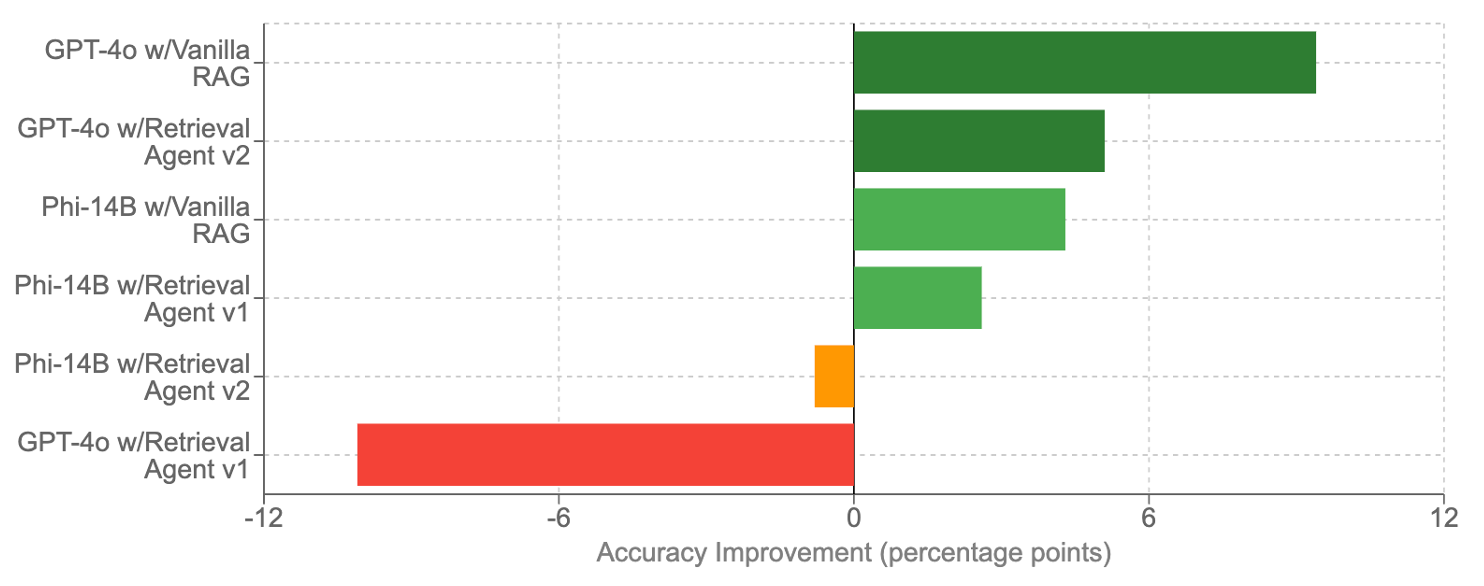}
\end{center}
\caption{Performance gains  (accuracy in percentage points) over vanilla LLMs (GPT4o and Phi4) on BioSage cross-domain benchmark. }
\label{fig:cross}
\end{wrapfigure}
GPT-4o with Vanilla RAG showed the largest improvement (+9.4 percentage points, +11.1\%) over the base model, demonstrating the value of retrieval-augmented generation for cross-domain tasks as demonstrated in Figure~\ref{fig:cross}. Phi-14B showed more consistent improvements across different system configurations, with both vanilla RAG and retrieval agent v1 outperforming the base model. For GPT-4o, we observed that retrieval agent v1 significantly underperformed compared to the base model, while retrieval agent v2 showed improvements. This suggests that our architectural refinements between agent versions successfully addressed limitations in the initial implementation. Across both GPT-4o and Phi-14B, vanilla RAG  outperformed agent-based approaches, indicating that for cross-disciplinary questions, the simpler retrieval approach may be more effective than more complex agent orchestration.

\paragraph{Compound AI Ablation Study Analysis}
Our ablation study revealed that each component of the BioSage architecture contributes differently to overall performance. For example, {\bf Vanilla RAG} consistently provided substantial improvements over base LLMs across all benchmarks, with performance gains ranging from 1.9 to 9.4 percentage points. {\bf Retrieval Agent v1} with query planning showed mixed results, with significant improvements for Llama-3-70B but performance degradation for GPT-4o on certain benchmarks. This suggests that model-specific optimization is crucial for effective agent implementation. {\bf Retrieval Agent v2} (note, it was only used to evaluate on our cross-disciplinary benchmark) addressed limitations of v1, showing more consistent improvements across models. For GPT-4o, agent v2 improved performance by 5.0 percentage points on our cross-disciplinary benchmark compared to the base LLM. {\bf Base LLM selection} significantly influenced performance across science benchmarks. While GPT-4o generally outperformed Llama-3-70B and Phi-14B as a base model, the magnitude of improvement from our compound AI components varied substantially between models.

\begin{figure}[t!]
\begin{center}
\includegraphics[width=14cm]{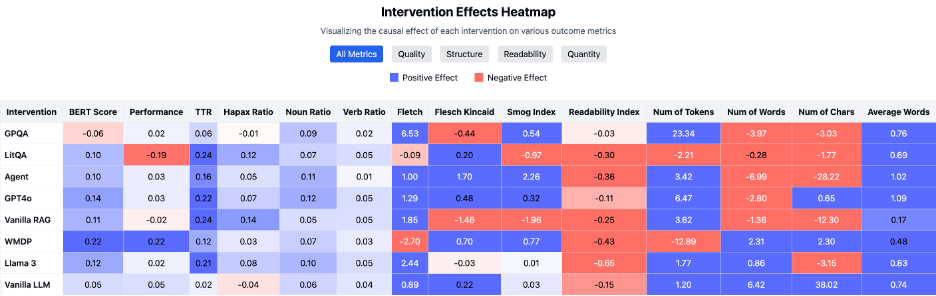}
\end{center}
\caption{Intervention effects heatmap visualizing the causal effect of each intervention on various outcome metrics across all tested configurations. Colors represent positive effects (blue) and negative effects (red) with intensity indicating effect magnitude. The heatmap reveals distinct causal patterns between model architectures, compound AI components, and performance outcomes across metrics.}
\label{fig:causal}
\end{figure}

\paragraph{Causal Investigations into Compound AI Behavior}
Our causal analysis reveals distinct patterns in how different BioSage components affect performance across evaluation metrics. By applying causal structure learning, we identified significant relationships between model architectures, compound AI components, and performance outcomes as visualized in Figure~\ref{fig:causal}. The intervention effects heatmap demonstrates that our agent-based implementations produce distinctly different causal patterns compared to vanilla LLMs and RAG-only approaches. Specifically, BioSage's retrieval agent demonstrates strong positive effects on structural metrics (Type-Token Ratio (TTR): +0.16, Noun Ratio: +0.11) and significantly improves performance on key benchmarks (Smog Index: +2.26). This indicates that our specialized agent architecture is particularly effective at improving document structure interpretation and complexity handling compared to baseline approaches.

The GPT4o-based implementation shows notable positive effects across linguistic complexity metrics (Flesch-Kincaid: +0.48) while maintaining strong performance on token efficiency (+8.47). Interestingly, the causal analysis reveals that vanilla RAG implementations have mixed effects, with negative impacts on performance (-0.02) despite improvements in TTR (+0.24). This supports our architectural decision to move beyond simple RAG toward more sophisticated agent orchestration. The most pronounced treatment effects appear in the WMDP benchmark, which shows exceptional performance improvements (+0.22) and readability scores (+0.77), suggesting that our compound AI architecture is particularly well-suited for biomedical knowledge domains. Contrastingly, the LitQA2 implementation demonstrates stronger effects on linguistic metrics (TTR: +0.34) but more modest performance gains (+0.10).

This causal analysis provides valuable insights for future iterations of BioSage. The negative effects observed in certain combinations (e.g., Vanilla LLM's poor performance on numeric metrics: -38.02 chars, -6.42 words) highlight the importance of our compound AI approach. The positive treatment effects of our agent implementations support our architectural decisions while pointing toward opportunities for further optimization in specific knowledge domains.

\section{Conclusions}
This paper presented BioSage, a novel compound AI architecture for cross-disciplinary scientific knowledge discovery and synthesis. Our comprehensive evaluation, including causal analysis and the introduction of a novel cross-disciplinary benchmark demonstrated significant performance improvements across multiple scientific benchmarks as well as performance explanation, with our agents outperforming vanilla RAG approaches by 13-21\% using GPT-4o and Llama-3 70B models. The results validate our approach to orchestrating specialized agents with frontier models and tools to bridge traditionally siloed domains. 

Our future work will focus on three key directions. First, we will enhance BioSage's multimodal retrieval capabilities to process scientific charts and tables, enabling deeper understanding of visual scientific content~\cite{huang2023lvlms,huang2024pixels,zhou2023enhanced}. Second, we plan to conduct extensive user studies with domain scientists to evaluate all three specialized agents (retrieval, translation, and reasoning) within our user-centric HAI workflows inspired by~\cite{volkova2025virtlab} for summarization, research debate, and cross-disciplinary brainstorming. Finally, we will develop a comprehensive multimodal cross-disciplinary benchmark specifically designed to evaluate compound AI systems' abilities to synthesize knowledge across both modalities and scientific domains. 

\newpage
\bibliography{references}
\bibliographystyle{neurips_2025}

\appendix
\newpage
\section{Technical Appendices and Supplementary Material}\label{appendix}
\subsection{Retrieval Agent Architecture and Implementation Details}
Retrieval agent incorporates self-reflection capabilities~\cite{huang2022large,pan2023automatically,shinn2023reflexion} to find and synthesize relevant knowledge across disciplines. Unlike conventional search algorithms, our retrieval agents employ query planning and synthesis strategies that significantly outperform vanilla LLM and RAG approaches~\cite{laurent2024labbenchmeasuringcapabilitieslanguage}. These agents not only locate relevant information but also consolidate findings across multiple sources, providing citation-backed responses that maintain scientific rigor.

\begin{figure}[htb!]
\begin{center}
\includegraphics[width=8.5cm]{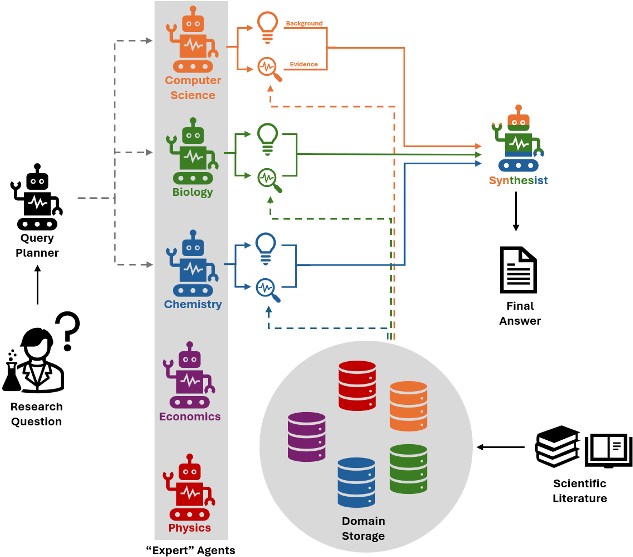}
\end{center}
\caption{Retrieval agent architecture illustrates the retrieval and synthesis workflow, which begins with a research question processed by a Query Planner that determines which domain-specific "Expert" RAG agents to engage (Computer Science, Biology etc.). Each domain RAG tool conducts parallel knowledge gathering through dual pathways—retrieving background knowledge and evidence from domain-specific storage repositories. All gathered information flows to a central Synthesist  that integrates cross-disciplinary insights to produce a comprehensive final answer.}
\label{fig:retrieval_agent}
\end{figure}

Figure~\ref{fig:retrieval_agent} demonstrates how the retrieval agent operates through a systematic five-step workflow that begins with query planning, where it identifies and prioritizes relevant knowledge domains for the user's question. It then performs domain evidence gathering by searching specialized corpora within each identified domain, collecting and summarizing pertinent information. Simultaneously, the agent evaluates text content from retrieved sources during the text selection phase, choosing supportive text that enhance understanding. To fill knowledge gaps, it conducts latent knowledge gathering by probing domain-specific models for contextual background information not explicitly stated in documents. Finally, in the synthesis phase, the agent integrates all collected evidence and domain perspectives into a comprehensive, multi-faceted response that bridges disciplinary boundaries and provides holistic insights that would be difficult to achieve through single-domain approaches.

\begin{tcolorbox} 
[colback=gray!5,colframe=black!75,title=Plan Query v1 Prompt,breakable]
\parskip=5pt
\tiny
\textbf{System prompt}: "You are a helpful assistant. You will follow instructions and respond in JSON format"

\textbf{Prompt}: 
"""I am going to provide you with a list of topic tags that correspond to different data sources, and a query. Your job is to identify which of those tags are most relevant to the query. Meaning, which data sources are most likely to contain the information that can provide good context to answer the query. You may choose as many tags as you want, or only one. Present selected tags in a list, ordered from most relevant to least relevant. If the question explicitly requests focus on one tag, return only that tag.

Tags: \{category\_tags\}

Question: \{prompt\}

Your response:
\{\{tags:[<fill in>, <fill in>, <fill in>, ...]\}\}"""
\end{tcolorbox}

\begin{tcolorbox} 
[colback=gray!5,colframe=black!75,title=Plan Query v2 Prompt,breakable]
\parskip=5pt
\tiny
\textbf{System prompt}: "You are a helpful assistant. You will follow instructions and respond in JSON format"

\textbf{Prompt}: 

"""I am going to provide you with a question, and I need you to convert it into a list of searchable terms that will help surface relevant literature in my database. Based on the given question come up with a list of terms that should generate a high response in a TF-IDF based search algorithm. Select search terms that are high in specificity, knowing that there is guaranteed to be at least one document in the database that pertains directly to the question, and it must be recovered with precision. You may generate as many search terms as you wish. Provide terms in a list, ordered from most relevant to least relevant.

Question: 
\{prompt\}

Your response: \{\{keywords: [<fill in>, <fill in>, <fill in>, ...]\}\}"""
\end{tcolorbox}

\begin{tcolorbox} 
[colback=gray!5,colframe=black!75,title=Evidentiary Expertise Prompt,breakable]
\parskip=5pt
\tiny
\textbf{System prompt}: "Answer in a direct and concise tone, I am in a hurry. Your audience is an expert, so be highly specific. If there are ambiguous terms or acronyms, first define them. Do not editorialize or explain the nature of your task."

\textbf{Prompt}: 
"""A {tag} related question has been asked by a researcher, and you are an expert in {tag}. You have also found some evidence that might help answer the question by searching the literature, and summarized it in your notes. Your job is to help explain the releveant evidence, and provide background information to ground and contextualize this information within the scope of the original question. Your main goal is to give the researcher the tools necessary to answer the question for themselves - without directly answering the question. Constrain the scope of your answer to your field of expertise, providing a unique lens through which to approach the question.

Question: \{query\_prompt\}

Evidence: \{evidence\}"""
\end{tcolorbox}

\begin{tcolorbox} 
[colback=gray!5,colframe=black!75,title=Perspective Sythesis Prompt,breakable]
\parskip=5pt
\tiny
\textbf{System prompt}: "You are the head of the research department at Hypothetical University, leading a pilot program designed to help students navigate the overwhelming volume of modern research. Your role is to synthesize expert insights into well-reasoned, evidence-based answers.

Your team consists of experts from various fields who gather and summarize relevant research papers. You must critically evaluate their inputs, resolve contradictions, and ensure that only the most accurate and relevant information is shared. As department head, your responses must be rigorous, neutral, and structured for clarity.

Your audience consists of advanced students, so be highly specific. Do not hallucinate. Do not editorialize or explain your task. Provide responses in JSON FORMAT, ensuring clarity and structured presentation of findings.

\textbf{Prompt}: """A student has asked the following question:
\{query\_prompt\}

Experts from the most relevant fields were identified and consulted for their input. They have all provided direct evidence for their answers. Integrate their responses, resolving inconsistencies where necessary.
Expert input:

\{context\}
Provide both a concise answer and a more thorough explanation. If the question is multiple-choice, the concise answer should state only the correct choice. Respond strictly in JSON format using this structure:

\{\{ "answer": "<fill in>", "explanation": "<fill in>"\}\}"""
\end{tcolorbox}

\begin{tcolorbox} 
[colback=gray!5,colframe=black!75,title=Evidence RAG Template Prompt,breakable]
\parskip=5pt
\tiny
\textbf{System prompt}: "Answer in a precise and scholarly tone. Your audience is an expert, so be highly specific. Do not hallucinate. Do not editorialize or explain the nature of your task. Provide your responses in JSON format."

\textbf{Prompt}: 
"""Imagine you are a {tag} research analyst specializing in summarizing research articles for junior scholars. Your job is to determine whether a retrieved journal article contains useful evidence for answering a question. 

Relevance is determined by whether the article provides **any** information that contributes to answering the question. This includes:

- Direct evidence that explicitly addresses the question.

- Background information that provides necessary context.

- Indirect evidence that helps formulate a reasoned answer. 

**If the article is relevant**, extract and summarize the evidence **accurately and comprehensively** while maintaining conciseness. Do not omit key details that could influence the final answer. 

Your response should be in a formal, report-style format, as it will be stored for future reference to improve retrieval methods. Follow this format: \{\{\{\{"relevant": <BOOL>, "summary": <optional>\}\}\}\}

This was the junior researcher’s question: \{\{query\_str\}\}

And this was the paper the librarian found: \{\{context\_str\}\}"""
\end{tcolorbox}

\newpage
\subsection{Translation Agent Architecture and Implementation Details}
Translation agents help bridge knowledge across different  domains e.g., biochemistry and AI. These agents address the fundamental challenge of cross-disciplinary communication by aligning specialized terminology, methodologies, and conceptual frameworks between fields. By serving as intermediaries between domain-specific knowledge bases, translation agents enable researchers from different disciplines to effectively communicate complex concepts without requiring expertise in each other's specialized vocabularies.

\begin{figure}[htb!]
\centering
\begin{subfigure}{.85\textwidth}
  \centering
  \includegraphics[width=.95\linewidth]{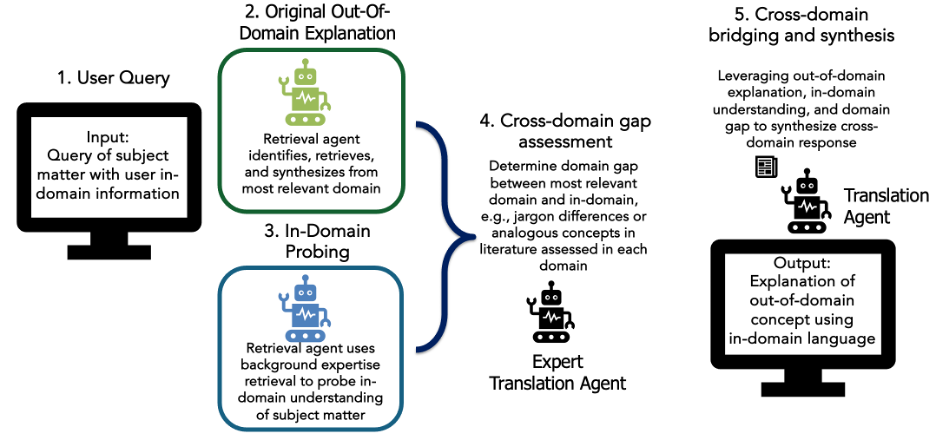}
  \caption{Persistent agent memory}
  \label{fig:litqa}
\end{subfigure}\\
\vspace{0.3cm}
\begin{subfigure}{.85\textwidth}
  \centering
  \includegraphics[width=.95\linewidth]{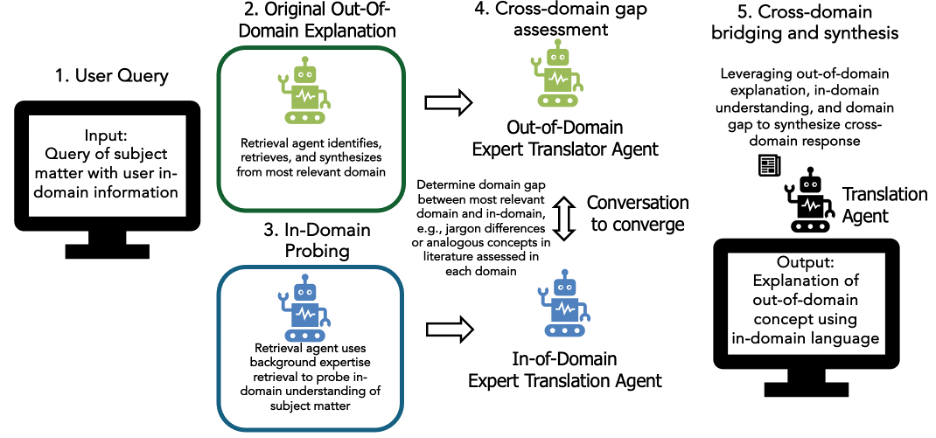}
  \caption{Shared cross-domain awareness via interactions.}
\end{subfigure}
\caption{Translation agent architecture variations for two implementations of BioSage's translation agent: (a) Persistent agent memory approach, where a single translation agent maintains knowledge across domains through internal memory representations, and (b) Shared cross-domain awareness via interactions, where specialized out-of-domain and in-domain translator agents engage in explicit conversation to bridge knowledge gaps.}
\label{fig:translation}
\end{figure}

As Figure\ref{fig:translation} demonstrates, our translation workflows operate through a structured five-step process designed to bridge these disciplinary divides. Beginning with a user query containing in-domain information, the system first engages retrieval agents to generate explanations from the most relevant domain (out-of-domain perspective). Simultaneously, it probes the user's own domain (in-domain) to establish contextual understanding of the subject matter. In the critical gap assessment phase, the translation agent identifies specific terminology differences, conceptual analogies, and methodological variations between domains by analyzing data from both perspectives. Finally, the synthesis step leverages this cross-domain understanding to produce explanations that effectively translate complex concepts into the user's domain-specific language. We implemented two architectural variations: a persistent memory approach that maintains cross-domain knowledge within a single agent, and a multi-agent conversational approach where domain-specific translation experts engage in explicit dialogue to negotiate meaning across disciplinary boundaries.

\begin{tcolorbox} 
[colback=gray!5,colframe=black!75,title=Gap Assessment,breakable]
\parskip=5pt
\tiny
\textbf{System prompt}: "Answer as correctly as possible with the provided information. Your audience is an expert, so be highly specific. Do not hallucinate. Do not editorialize or explain the nature of your task."

\textbf{Prompt}: """What is the knowledge gap between the following answers and fields of \{out\_of\_domain\_tags\_str\}, and \{in\_domain\_tags\_str\} from the question \{orig\_prompt\}?

A \{out\_of\_domain\_tags\_str\} answer for the question \{orig\_prompt\} is as follows: \{inputs.out\_of\_domain\_answer\}

A \{in\_domain\_tags\_str\} answer for the question \{orig\_prompt\} is as follows: \{inputs.in\_domain\_answer\}"""
\end{tcolorbox}

\begin{tcolorbox}[colback=gray!5,colframe=black!75,title=Gap Bridge,breakable]
\parskip=5pt
\tiny
\textbf{System prompt}: "Answer as correctly as possible with the provided information. Your audience is an expert, so be highly specific. Do not hallucinate. Do not editorialize or explain the nature of your task."

\textbf{Prompt}:
"""Improve the \{in\_domain\_tags\_str\} answer below from the question \{orig\_prompt\} using the knowledge from \{out\_of\_domain\_tags\_str\}.
Cater the new answer and draw parallels to a \{in\_domain\_tags\_str\} experts, such that they would understand it.\\
Output the new answer.\\
Note that there exists a knowledge gap between these domains with the summary of:
\{inputs.knowledge\_gap\}

A \{out\_of\_domain\_tags\_str\} answer for the question \{orig\_prompt\} is as follows: \{inputs.out\_of\_domain\_answer\}

A \{in\_domain\_tags\_str\} answer for the question \{orig\_prompt\} is as follows: \{inputs.in\_domain\_answer\}"""
\end{tcolorbox}

\begin{tcolorbox}
[colback=gray!5,colframe=black!75,title=Background Expertise,breakable]
\parskip=5pt
\tiny
\textbf{System prompt}: "Answer in a direct and concise tone, I am in a hurry. Your audience is an expert, so be highly specific. If there are ambiguous terms or acronyms, first define them. Do not editorialize or explain the nature of your task."

\textbf{Prompt}: """A \{inputs.discipline\} related question has been asked by a researcher, and you are an expert in \{inputs.discipline\}. Provide the best evidence you can from your background knowledge in the topic to give the researcher the information necessary to answer the question for themselves - without directly answering the question. Constrain the scope of your answer to your field of expertise, providing a unique lens through which to approach the question.

        Question: \{orig\_prompt\}"""

\end{tcolorbox}

\newpage
\subsection{Reasoning Agent Architecture and Implementation Details}
\begin{figure}[htb]
\begin{center}
\includegraphics[width=5.5cm]{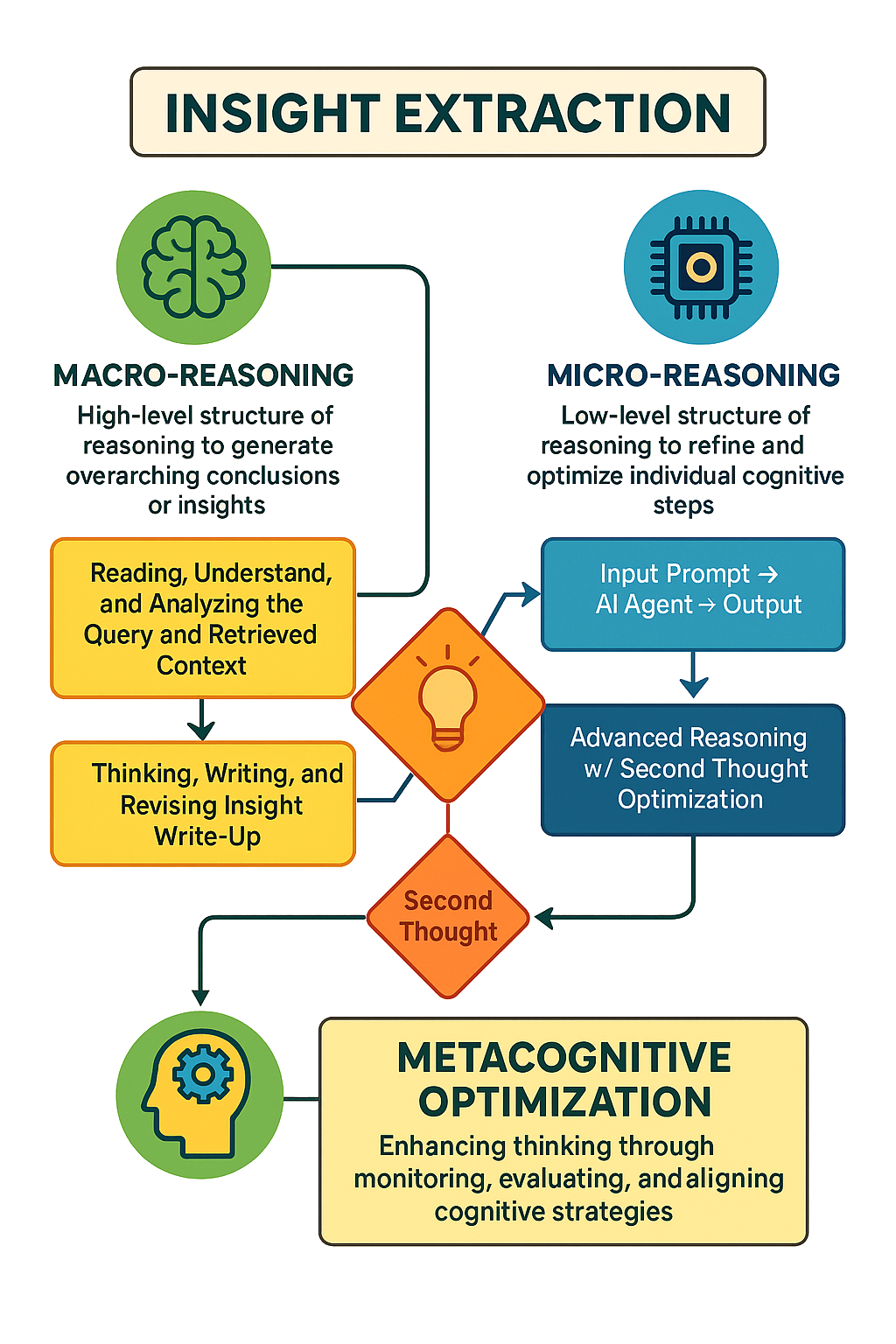}
\end{center}
\caption{Multi-level reasoning workflows in BioSage's insight extraction. BioSage's three-tiered reasoning architecture for scientific knowledge synthesis. Macro-reasoning focuses on high-level structural reasoning to generate overarching conclusions through reading, understanding, and analyzing query context, followed by thinking, writing, and revising insights. Micro-reasoning addresses low-level reasoning structure to refine and optimize individual cognitive steps through a sequential process from input prompt to output, enhanced by advanced reasoning with second thought optimization. At the bottom, metacognitive optimization represents the self-reflective layer that enhances thinking by monitoring, evaluating, and aligning cognitive strategies.}
\label{fig:reasoning}
\end{figure}

Reasoning agents extend retrieval agents that perform knowledge discovery and synthesis across domains by employing both micro-reasoning for detailed analysis of specific evidence and macro-reasoning for integrating diverse insights~\cite{dedhe2023origins,koskinen2013macro} as describe in Figure~\ref{fig:reasoning}. Our reasoning agents incorporate advanced memory mechanisms spanning semantic, episodic, and procedural structures~\cite{squire2004memory,zhang2024survey} that enable them to maintain context, learn from interactions, and adapt their behaviors across scientific workflows. 

The agent employs second-thought processes~\cite{stone2022second,stanovich2020humans,pennycook2015makes,mauboussin2012think,evans2010intuition} for self-correction and meta-cognitive optimization strategies~\cite{hough2019understanding,fleming2012metacognition,fletcher2012metacognition,coutinho2008self,clark1988metacognition} to continuously improve performance. The agent architecture is deliberately designed to provide transparent explanations of reasoning processes, enabling users to understand how conclusions are reached and maintain appropriate trust calibration.

\newpage
\subsection{BioSage Agent Architecture}
BioSage's agent architecture represents a significant advancement in implementing cognitive mechanisms for scientific knowledge discovery and synthesis. Unlike traditional AI systems, our agents employ a sophisticated cognitive architecture modeled after human expert reasoning patterns. As illustrated in Figure~\ref{fig:reasoning}, each agent maintains a comprehensive profile that includes domain expertise, operational goals, and interaction priorities. At the core of this architecture is the behaviors module, which implements a continuous decision loop that processes input prompts, evaluates situational context, applies appropriate guardrails, and selects actions or tool invocations. This decision-making process is informed by a multi-level memory system that distinguishes between semantic knowledge (domain facts and relationships), procedural knowledge (methods and techniques), and episodic memory (prior interactions and experiences). This memory differentiation enables agents to not only retrieve relevant information but also apply appropriate reasoning patterns based on context. The external connections to world state, observations, and RAG systems allow agents to ground their reasoning in both retrieved literature and ongoing interaction context. This architecture enables BioSage agents to mimic human expert reasoning strategies when navigating complex cross-disciplinary knowledge spaces, providing users with insights that effectively bridge traditionally siloed domains while maintaining scientific rigor and transparency.

\begin{figure}[htb!]
\begin{center}
\includegraphics[width=10cm]{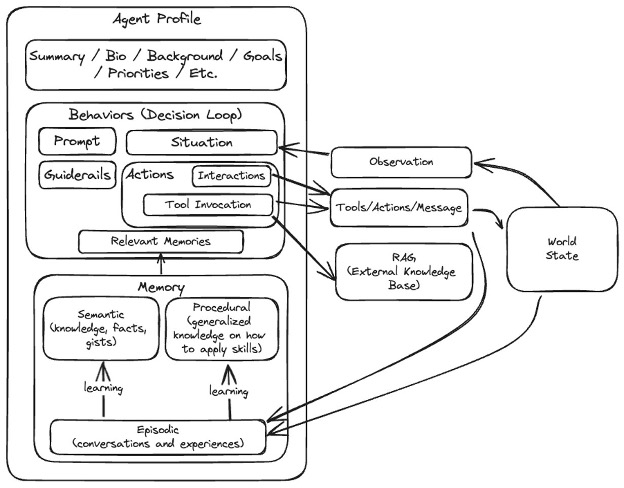}
\end{center}
\caption{BioSage cognitive language agent architecture featuring layered components that enable sophisticated cross-disciplinary reasoning. The architecture includes a profile  with summary information and goals, a central behaviors module with decision loops for processing prompts, evaluating situations, and executing actions within guardrails, and a multi-faceted memory system that integrates semantic knowledge (facts), procedural knowledge (skills), and episodic memory (conversation history). External components include connections to observation capabilities, tool invocation mechanisms, RAG for external knowledge retrieval, and world state modeling.}
\label{fig:architecture}
\end{figure}

\newpage
\subsection{Detailed Comparative Analysis on LitQA2 Benchmark}\label{appendix:eval}

\begin{figure}[htb]
\begin{center}
\includegraphics[width=10cm]{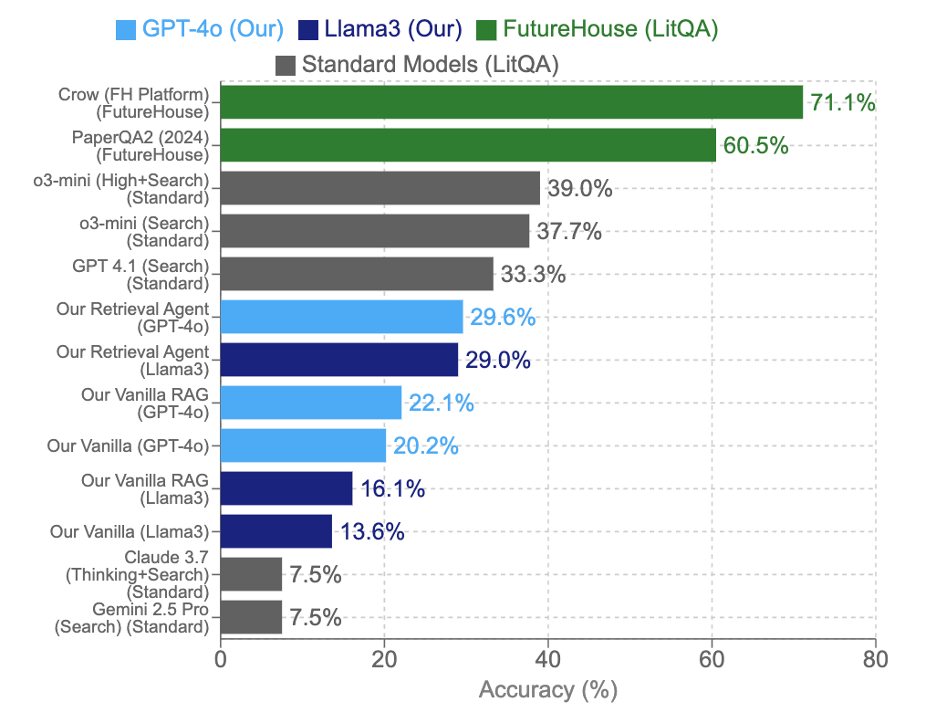}
\end{center}
\caption{Accuracy comparison of various LLMs and configurations on the LitQA2 benchmark. FutureHouse's specialized scientific agents (green) achieve the highest accuracy, with Crow at 71.1\% and PaperQA2 at 60.5\%. Other LLMs (gray) show moderate performance (33.3\%-39.0\%), while our BioSage retrieval agent implementations (blue/navy) demonstrate substantial improvement over vanilla configurations, achieving 29.6\% and 29.0\% with GPT-4o and Llama 3.1 base respectively.}
\label{fig:other}
\end{figure}

For the LitQA2 benchmark comparison,\footnote{https://www.futurehouse.org/research-announcements/launching-futurehouse-platform-ai-agents} BioSage's retrieval agents (29.6\% for GPT-4o and 29.0\% for Llama-3) demonstrated competitive performance against other systems like Elicit (25.9\%), significantly outperforming vanilla models as shown in Figure~\ref{fig:other}. Only specialized scientific retrieval systems like FutureHouse's Crow (71.1\%) and PaperQA2 (60.5\%) achieved substantially higher accuracy, highlighting the effectiveness of domain-specific optimization.

Note, BioSage offers distinct advantages through its specialized cross-disciplinary approach. Unlike FutureHouse which focuses primarily on within-domain scientific discovery with agents like Crow (concise search), Falcon (deep literature reviews), Owl (precedent search), and Phoenix (molecular synthesis), BioSage is specifically engineered for cross-disciplinary knowledge discovery and synthesis across multiple scientific domains.
BioSage's architecture uniquely combines RAG with foundation models, specialized agents and tools to enable breakthrough connections between traditionally siloed domains. Its strength lies in specialized scientific workflows—including cross-disciplinary translation, expert consultation, research debate, and brainstorming—that bridge communication gaps between scientists from different fields. While both systems employ agent architectures, BioSage's agents are specifically designed to reason across disciplinary boundaries.

While benchmark evaluations like LitQA2 provide important quantitative comparisons, BioSage's distinctive focus on cross-disciplinary workflows necessitates evaluation across multiple specialized benchmarks including GPQA, WMDP, and HLE-Bio to fully capture its unique capabilities. Beyond computational metrics, BioSage emphasizes human-AI interaction (HAI) evaluation through structured usability studies with domain scientists across different fields, directly assessing how effectively the system enhances cross-disciplinary collaboration and knowledge discovery in practice. This dual evaluation approach—combining diverse technical benchmarks with real-world collaboration assessments—provides a more complete picture of BioSage's value proposition compared to FutureHouse's primarily literature-focused performance metrics. By measuring both technical performance and practical scientific workflow enhancement, BioSage establishes a more comprehensive evaluation framework aligned with its core mission of bridging disciplinary knowledge gaps in ways that quantitative benchmarks alone cannot capture.

\end{document}